\documentclass[10pt,twocolumn,letterpaper]{article}

\usepackage{cvpr}              
\usepackage{multirow}
\usepackage{booktabs}
\usepackage{bbding}
\usepackage{amssymb}
\usepackage{fontawesome5}
\usepackage{xcolor}

\definecolor{mygold}{HTML}{FFD700}
\definecolor{cvprblue}{rgb}{0.21,0.49,0.74}
\usepackage[pagebackref,breaklinks,colorlinks,allcolors=cvprblue]{hyperref}

\def\confName{CVPR}
\def\confYear{2026}

\newcommand{\decrease}[1]{\textcolor{red}{\footnotesize(#1)}}

\title{Towards Real-World Document Parsing via Realistic Scene Synthesis and Document-Aware Training}

\author{
Gengluo Li$^{1,4,\dagger}$ \quad
Pengyuan Lyu$^{2,\dagger}$ \quad
Chengquan Zhang$^{2,\ddagger}$ \quad
Huawen Shen$^{1,4}$ \quad
Liang Wu$^{2}$ \\
Xingyu Wan$^{2}$ \quad
Gangyan Zeng$^{5}$\textsuperscript{\scalebox{0.8}{\faEnvelope}} \quad
Han Hu$^{2}$ \quad
Can Ma$^{1,4}$ \quad
Yu Zhou$^{3}$\textsuperscript{\scalebox{0.8}{\faEnvelope}} \\[0.25em]
$^{1}$Institute of Information Engineering, Chinese Academy of Sciences \quad
$^{2}$Tencent \\
$^{3}$Nankai University \quad
$^{4}$University of Chinese Academy of Sciences \\
$^{5}$Nanjing University of Science and Technology \\[0.25em]
{\tt\small ligengluo@iie.ac.cn \quad yzhou@nankai.edu.cn \quad gyzeng@njust.edu.cn} \\
{\small $^{\dagger}$Equal contribution \quad
$^{\ddagger}$ Project leader \quad
\textsuperscript{\scalebox{0.9}{\faEnvelope}} Corresponding author}
}

\begin{document}
\maketitle
\begin{abstract}
Document parsing has recently advanced with multimodal large language models (MLLMs) that directly map document images to structured outputs. Traditional cascaded pipelines depend on precise layout analysis and often fail under casually captured or non-standard conditions. Although end-to-end approaches mitigate this dependency, they still exhibit repetitive, hallucinated, and structurally inconsistent predictions—primarily due to the scarcity of large-scale, high-quality full-page (document-level) end-to-end parsing data and the lack of structure-aware training strategies. To address these challenges, we propose a data–training co-design framework for robust end-to-end document parsing. A Realistic Scene Synthesis strategy constructs large-scale, structurally diverse full-page end-to-end supervision by composing layout templates with rich document elements, while a Document-Aware Training Recipe introduces progressive learning and structure-token optimization to enhance structural fidelity and decoding stability. We further build Wild-OmniDocBench, a benchmark derived from real-world captured documents for robustness evaluation. Integrated into a 1B-parameter MLLM, our method achieves superior accuracy and robustness across both scanned/digital and real-world captured scenarios. All models, data synthesis pipelines, and benchmarks will be publicly released to advance future research in document understanding.
\end{abstract}    
\section{Introduction}
\label{sec:intro}

\begin{figure}[ht]
    \centering
    \includegraphics[width=1\linewidth]{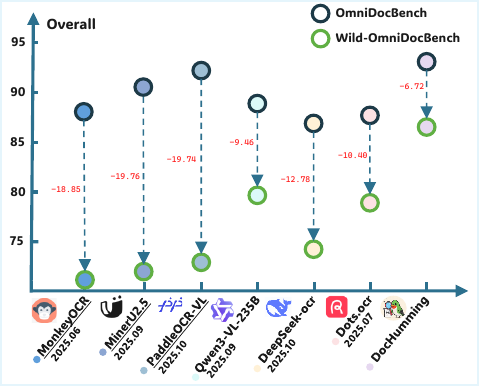}
    \vspace{-10pt}
    \caption{Overall Performance and Degradation from OmniDocBench to Wild-OmniDocBench. \underline{Underlined} method names correspond to modular cascaded pipelines.}
    \vspace{-10pt}
    \label{fig:leida}
\end{figure}

Documents serve as vital information carriers across domains, from historical manuscripts to academic papers and business contracts. 
As large language models (LLMs) advance, reliable document parsing becomes increasingly crucial~\cite{rag_storm,rag_doc}, enabling the transformation of unstructured visual inputs into structured, machine-readable outputs for tasks such as digitization, information access, and workflow automation~\cite{mplug_doc}.

\begin{figure*}[ht]
    \centering
    \includegraphics[width=1\linewidth]{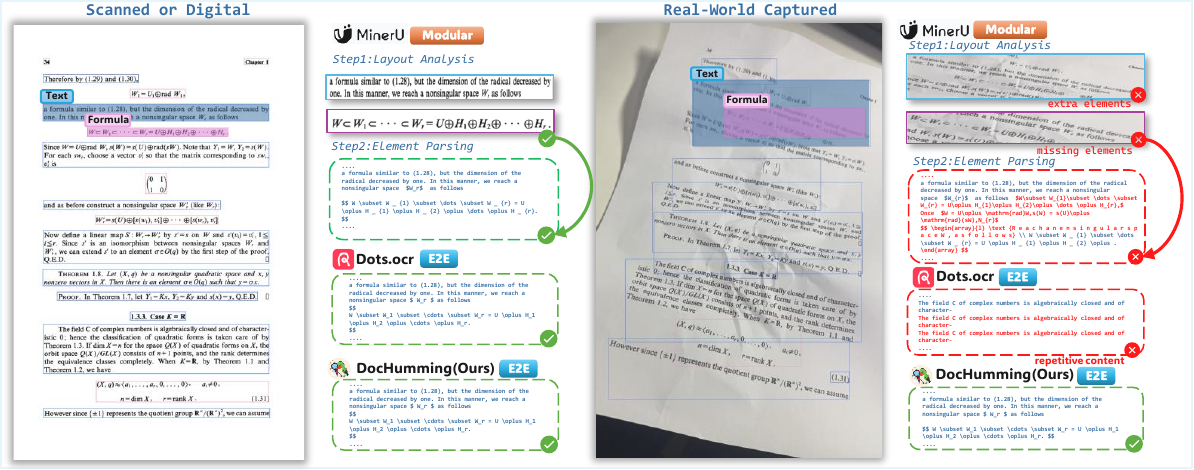}
    \vspace{-10pt}
    \caption{\textbf{Scanned/Digital and Real-World Capture.} On scanned/digital pages, both modular and E2E parsers decode correctly. Under real-world capture, modular cascades accumulate layout-analysis errors that propagate to element parsing (extra/missing regions), while generic end-to-end models exhibit repetitive outputs.}
    \vspace{-10pt}
    \label{fig:shoutu}
\end{figure*}

Document parsing has evolved from early modular systems~\cite{mineru}, where tasks such as text spotting~\cite{text_recog_chen1,jiahao_spot}, layout analysis~\cite{layoutyolo,Huawen_layout} were handled by separate components, followed by content-specific parsing modules~\cite{text_recog_chen2,text_recog_chen3,zeng_form,xiaomeng_recog} (e.g., for tables or forms). 
While this cascaded design enabled targeted processing, it was susceptible to error propagation, required manual coordination, and lacked generalization to diverse document types. 
Recent advances in multimodal large language models (MLLMs) have enabled end-to-end parsing by directly mapping document images to structured outputs, effectively by passing layout-based decomposition and cascading pipelines~\cite{mllm_got,dots.ocr2025,wei2025deepseek}. 
However, despite their success on digital-born documents, these models often struggle in real-world scenarios—producing repetitive content, hallucinations when handling casually captured or scanned layouts (Figure~\ref{fig:shoutu}). 
These deficiencies largely stem from the scarcity and high cost of large-scale, high-quality data for end-to-end document parsing.

To alleviate the cost of large-scale end-to-end supervision, recent works have revisited cascade designs that decompose document parsing into successive sub-tasks. 
A layout model or layout-prompted MLLM first segments the page, and the resulting sub-images are then parsed individually and aggregated into a unified structured representation.
This divide-and-conquer paradigm leverages abundant layout/element data and can surpass end-to-end models on \emph{scanned or digital} settings, as reflected by results on OmniDocBench\cite{ominibench} (see Fig.~\ref{fig:leida}).

However, this return to cascaded pipelines—though reducing reliance on large-scale end-to-end parsing data—remains heuristic and inherits the intrinsic limitations of modular systems, particularly the dependence on accurate layout segmentation.
As illustrated in Fig~\ref{fig:shoutu}, early-stage errors readily propagate downstream, especially on non-standard or casually captured documents.
Moreover, such task-specific modularity limits the development of general-purpose MLLMs with holistic perception and high-level reasoning over complex documents, constraining their potential for comprehensive document understanding.
These observations motivate us to revisit the necessity and potential of a truly end-to-end paradigm for document parsing.

We identify three key objectives for advancing end-to-end document parsing:  
(i) \textbf{Data scalability.} The scaling law of end-to-end document parsing remains insufficiently validated due to the lack of automated pipelines for generating large-scale, high-quality end-to-end parsing data, leaving the data barrier unresolved.  
(ii) \textbf{Task-specific training strategies.} Current approaches largely inherit standard MLLM training procedures without adaptation to the structural complexity and contextual dependencies of document parsing tasks.  
(iii) \textbf{Robustness in real-world scenarios.} Cascaded pipelines rely on precise layout analysis, which becomes unreliable under non-standard or casually captured conditions. In contrast, end-to-end designs, free from explicit layout dependency, hold greater potential for robust document understanding in such settings.

To address these objectives, we propose a \textbf{Realistic Scene Synthesis} framework at the data level, which combines diverse layout templates with a rich element repository to generate scalable end-to-end parsing data—including synthetic samples with explicit reading order, structural diversity, and data augmentation pipelines. 
At the training level, we introduce a \textbf{progressive learning strategy} inspired by the short-to-long context curriculum in LLM training.
The model first learns to parse isolated elements, then gradually transitions to full-document inputs for unified long-context understanding.
To further enhance structural awareness and decoding stability, we apply a \textbf{structure-token-aware optimization} that emphasizes structurally critical tokens.
Together, these techniques constitute our \textbf{Document-Aware Training Recipe}.
Finally, for evaluation, we construct a benchmark adapted from scanned and digital document datasets into \textit{real-world captured} styles, enabling assessment under authentic wild scenarios and providing new insights into the robustness of document parsing systems.

\noindent Our main contributions are summarized as follows:
\begin{itemize}
    \item \textbf{Realistic Scene Synthesis.} 
    We propose a scalable data construction framework that unifies fine-grained document elements with diverse layout templates, supporting end-to-end parsing with structural diversity and multilingual coverage. 

    \item \textbf{Document-Aware Training Recipe.} 
    We propose a \textbf{progressive learning strategy} and a \textbf{structure-token-aware optimization} to enhance structured parsing. 

    \item \textbf{Wild-OmniDocBench Benchmark.} 
    We construct a new evaluation benchmark tailored to \textit{real-world captured} document scenarios, providing a comprehensive assessment of parsing robustness under wild conditions.
\end{itemize}

\section{Related Work}
\label{sec:formatting}

\noindent\textbf{Modular Paradigm.}  
Traditional document parsing systems adopt a modular pipeline, where layout analysis~\cite{layouttrans2,layoutllm} segments document elements that are then processed by specialized modules for text extraction~\cite{Text_2,Text_3}, table recognition~\cite{table_1,table_2}, and formula parsing~\cite{form_1,form_2}.  
Each component is optimized independently, and the outputs are integrated to form the final structured representation.  
Recent MLLM-based approaches~\cite{li2025monkeyocr,feng2025dolphin} retain this pipeline by parsing segmented regions with heterogeneous prompts under a unified model.  
While leveraging MLLMs’ generalization ability, these methods still rely on layout segmentation and fail to enhance holistic document perception.

\noindent\textbf{End-to-End Paradigm.}  
End-to-End (E2E) systems treat document parsing as a sequence-to-sequence task that maps document images directly to structured text.  
Recent advances\cite{mllm_got,mllm_vary,dots.ocr2025,mplug_doc, team2025hunyuanocr} unify text, table, and formula parsing within a shared decoding framework, achieving strong performance on clean, digital-born documents via large-scale vision–language pretraining and PDF-to-LaTeX supervision.  
However, E2E models still struggle in real-world scenarios with complex or casually captured layouts, often producing repetitive, missing, or structurally inconsistent outputs.  
Their robustness and generalization remain limited by two main factors:

\begin{itemize}
\item \textbf{Data limitations.}  
While table and formula parsing benefit from task-specific datasets, large-scale and diverse data for unified E2E document parsing remain scarce.

\item \textbf{Optimization limitations.}  
Most models adopt uniform autoregressive decoding~\cite{llava}, overlooking the hierarchical structure of documents such as tables and forms.  
Without structure-aware objectives, E2E models tend to produce repetitive content and inconsistent layouts, particularly in long or layout-heavy documents.
\end{itemize}

These challenges indicate that the potential of E2E document parsing remains underexplored, motivating our data-centric synthesis and structure-aware training strategies.

\noindent\textbf{Data Engine.}  
Synthetic data generation is essential for scaling document parsing, yet existing engines remain limited in diversity and scope.  
\textit{SynthDog}\cite{synthdog} generates simple text layouts with minimal structural variation.  
\textit{DocLayout-YOLO}\cite{layoutyolo} offers diverse layout synthesis but focuses solely on layout analysis rather than full document parsing.  
\textit{GOT}\cite{mllm_got} produces digital-born data via PDF-to-LaTeX conversion, resulting in uniform layouts that lack the visual complexity of real-world captured scenes.  
These limitations underscore the need for a data engine that can generate end-to-end parsing datasets reflecting realistic structures and diverse visual conditions.

\noindent\textbf{Document Parsing Benchmarks.}  
Existing benchmarks mainly evaluate document parsing on scanned or digital data, lacking the imperfections of real-world captured documents.  
Fox~\cite{bench_fox} targets scanned Chinese and English documents but omits complex structures such as tables and formulas.  
OmniDocBench~\cite{ominibench} broadens structural coverage yet remains confined to clean, well-aligned digital pages, without modeling distortions, shadows, or illumination variations common in casually captured scenes.  
As a result, current benchmarks fail to represent the challenges encountered in real-world document parsing.

\begin{figure*}[ht]
    \centering
    \includegraphics[width=1\linewidth]{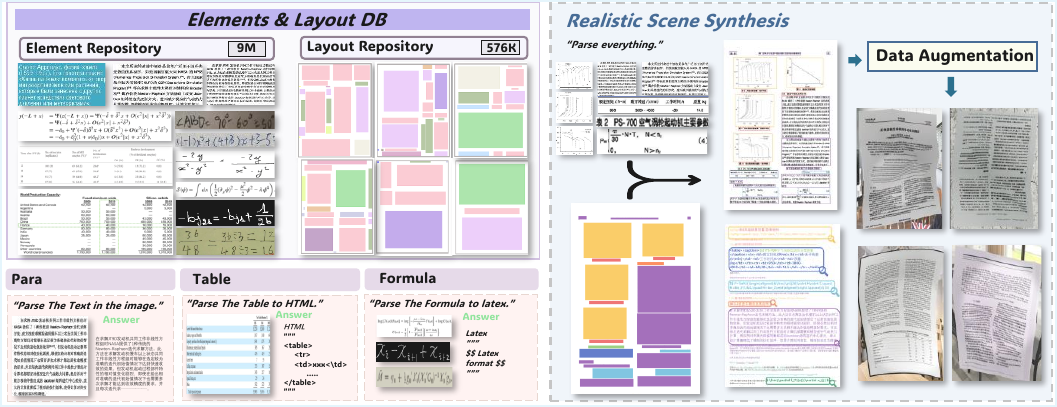}
    \vspace{-10pt}
    \caption{\textbf{Overview of Realistic Scene Synthesis.} Left: repositories of atomic elements and layout templates with reading order. Right: a synthesis pipeline that composes sampled elements into templates under spatial/structural constraints to produce page-level annotations, followed by capture-aware augmentation  to simulate real-world images.}
    \vspace{-10pt}
    \label{fig:Fig1}
\end{figure*}

\section{Realistic Scene Synthesis}
\label{sec:realistic_synthesis}

High-quality end-to-end document parsing data—balancing scale and diversity—remains scarce due to expensive annotation, whereas abundant resources exist for individual elements. 
We therefore develop a systematic and scalable synthesis framework that consolidates fine-grained element knowledge into large-scale, realistic page-level document data for end-to-end parsing.

\noindent\textbf{Pipeline Overview.}
As shown in Figure~\ref{fig:Fig1}, we generate realistic samples by composing atomic elements with curated layout templates. 
We first construct a repository of standardized elements (tables, formulas, paragraphs, figures) from multiple sources, and collect layout templates with annotated reading order to capture structural patterns observed in real documents. 
Document instances are synthesized by placing sampled elements into templates under spatial and structural constraints, yielding diverse, layout-rich pages for end-to-end parsing.

\noindent\textbf{Element Repository Construction.}
We integrate datasets for table recognition~\cite{data_fintabnet,data_pubtabnet}, formula parsing~\cite{data_unimernet,data_hme,data_crohme19}, and paragraph understanding~\cite{bench_fox,mllm_got}, followed by format normalization for cross-source consistency. 
Figures are obtained by segmenting visual regions from real pages with layout analysis models. 
To further expand diversity, we employ Qwen2.5-72B~\cite{qwen2.5} to rewrite and augment annotations—reorganizing tables, perturbing formula symbols, and creating hybrids (e.g., formulas embedded in tables or paragraphs). 
We also generate semantically coherent and multilingual paragraph groups to enhance contextual and linguistic coverage. 
All elements are rendered as images with paired parsing labels via a LaTeX-based pipeline~\cite{pdftex}.

\noindent\textbf{Layout Library Construction.}
We collect public layout datasets with reading-order annotations~\cite{doclaynet2022,cdla2024} and mine additional real-world layouts from the web, filtered by a layout detector~\cite{layoutyolo}. 
Underrepresented styles are supplemented by composing partial templates, resulting in a library of over 576K layout patterns covering a wide spectrum of structures.

\noindent\textbf{Data Augmentation.}
To improve robustness to real-world capture, we simulate natural variations~\cite{warping} including geometric (perspective shifts, bends, wrinkles), photometric (illumination and exposure changes), camera (random rotations), and environmental (realistic background overlays). 
This narrows the gap between synthetic and casually captured documents and enhances resilience to noise and deformation.

\noindent\textbf{Data Scale and DocMix-3M.}
Our framework integrates \(\sim\)9M atomic elements with 576K layout templates to produce \textbf{DocMix-3M}, \(\sim\)3M high-quality synthetic documents, of which \(\sim\)20\% are augmented using the above pipeline to mimic casually captured conditions. 
DocMix-3M exhibits rich structural diversity and visual variability, supporting large-scale end-to-end training across domains and acquisition settings.


\section{Document-Aware Training Recipe}

Auto-regressive models often face challenges when handling varying context lengths during training — a problem well-studied in LLMs optimization, where curricula typically transition from short to long contexts to ensure stable convergence~\cite{xiong2023effective, liu2025comprehensive}. This issue similarly arises in end-to-end document parsing, where a clear context gap exists between sub-element parsing (e.g., formulas, tables) and full-document understanding of text-rich images. To address convergence stability and learning efficacy in such heterogeneous settings, we propose the \textbf{Document-Aware Training Recipe}, which integrates a progressive training paradigm with a structure-token aware optimization.

\noindent\textbf{Progressive Training Paradigm.}
To fully leverage our constructed data and improve the stability and generalization of end-to-end document parsing, we introduce a progressive training paradigm consisting of two stages. This design is inspired by established practices in LLM pretraining—starting from short-context tasks and gradually transitioning to long-context understanding.

In the first stage, we train the model to parse individual elements (e.g., tables, formulas, paragraphs) using heterogeneous prompts over isolated element images. This localized supervision avoids contamination from unannotated visual noise—commonly present in public element datasets—and allows the model to acquire type-specific parsing capabilities in a controlled context. Furthermore, we extend the vocabulary with layout-specific structure tokens (e.g., \texttt{<table>}, \texttt{<tr>}) to better support the autoregressive decoding of structured outputs.

In the second stage, DocMix-3M serves as the primary corpus for full-document training. We incorporate 1M samples from Stage 1 to retain element-level capabilities, and employ a unified prompt format to facilitate cohesive end-to-end decoding across diverse parsing tasks.

Overall, this progressive design serves two key purposes: (i) it ensures training stability by aligning context complexity with learning stages, and (ii) it provides a structured path from heterogeneous sub-element parsing to unified full-document understanding—supporting better capability transfer and holistic document modeling.

\noindent\textbf{Structure-Token Aware Optimization.}
To improve the stability of structured output under the autoregressive decoding paradigm, we introduce a structure-token aware optimization strategy. While conventional training treats all output tokens equally, structured content such as tables is more sensitive to inconsistencies, and errors (e.g., repeated rows, misaligned cells) can propagate destructively. To mitigate this, we assign higher loss weights to structured tokens enclosed within tags like \texttt{<table>} and \texttt{</table>} to guide the model toward more precise generation. Formally, the loss becomes:

\begin{equation}
L_{\text{structured}} = -\sum_{t=1}^{T} \alpha_t y_t \log P(x_t | x_{<t})
\end{equation}
where:
\begin{equation}
\alpha_t =
\begin{cases}
\lambda, & \text{if } x_t \text{ is a structured token} \\
1,       & \text{otherwise}
\end{cases}
\end{equation}

The goal of this targeted adjustment is to improve structural consistency and reduce repetitive predictions, particularly in structured outputs such as tables and hierarchies. The method was validated on a 1B-parameter MLLM, resulting in the \textbf{DocHumming} model—named after the hummingbird, a small yet agile creature capable of hovering precisely while efficiently converting energy from its intake, symbolizing compactness, precision, and efficiency in document understanding.

\section{Wild-OmniDocBench}
\label{sec:wild_omnibench}

To assess real-world robustness, we construct \textbf{Wild-OmniDocBench}, a benchmark for end-to-end parsing on naturally captured documents.
Compared with scanned or digital-born data, these images exhibit illumination variation, wrinkles, reflections, and geometric distortions that challenge models trained on clean digital datasets.
Wild-OmniDocBench enables systematic robustness evaluation in realistic capture scenarios.

\begin{figure}[ht]
    \centering
    \includegraphics[width=1\linewidth]{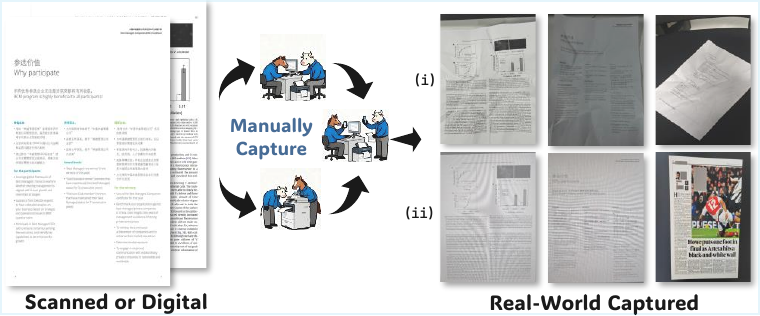}
    \vspace{-15pt}
    \caption{\textbf{Wild-OmniDocBench Construction.} We convert scanned pages into real-world–captured images by (i) printing, deforming, and photographing under varied lighting, and (ii) displaying on screens and re-shooting to induce moiré and reflections.}
    \vspace{-10pt}
    \label{fig:wildomni}
\end{figure}

\noindent\textbf{Construction Pipeline.}  
To create \textbf{Wild-OmniDocBench}, we manually convert the entire OmniDocBench~\cite{ominibench} into real-world–captured form, inspired by \cite{wilddoc}, via controlled acquisition and physical simulation, following the collection procedure illustrated in Fig.~\ref{fig:wildomni}.
Specifically, two complementary procedures are adopted.  
First, we print document pages and apply physical manipulations such as folding, bending, and crumpling to introduce realistic surface deformations.  
These printed documents are then photographed under diverse lighting conditions—including directional, uneven, and low-light setups—to simulate illumination changes observed in natural environments.  
Second, we display digital documents on various media devices such as computer monitors and smartphone screens, followed by photographic capture to emulate artifacts including moiré patterns, screen reflections, and brightness variations.

\section{Experiments}
\subsection{Datasets and Evaluation}

We adopt a progressive training paradigm built upon our Realistic Scene Synthesis pipeline. The model is evaluated on public and proposed benchmarks to assess generalization and robustness.

\begin{table*}[h]
    \centering
    \small
    \aboverulesep = 0.1ex
    \belowrulesep = 0.2ex
    \caption{Comparison of various OCR and VLM systems on document understanding benchmarks. Higher $ \uparrow $ indicates better performance, lower $ \downarrow $ indicates smaller error.}
    \vspace{-10pt}
    \begin{tabular}{clcccccc}
        \toprule
        \multirow{2}{*}{\textbf{Model$\mathrm{_{Type}}$}} & \multirow{2}{*}{\textbf{Method}} & \multirow{2}{*}{\textbf{Size}} & \multicolumn{5}{c}{\textbf{Metrics}} \\ 
        \cmidrule(l){4-8}
        & & & \textbf{Overall $\uparrow$} & \textbf{Text$^{\mathrm{Edit}}$ $\downarrow$} & \textbf{Formula$^{\mathrm{CDM}}$ $\uparrow$} & \textbf{Table$^{\mathrm{TEDS}}$ $\uparrow$} & \textbf{Reading Order$^{\mathrm{Edit}}$ $\downarrow$} \\
        \midrule

        \multirow{3}{*}{\parbox{1.6cm}{\centering \textbf{Pipeline\\Tools}}}
        & Marker-1.8.2\cite{marker2025} & - & 71.30 & 0.206 & 76.66 & 57.88 & 0.250 \\    
        & Mineru2-pipeline\cite{mineru} & - & 75.51 & 0.209 & 76.55 & 70.90 & 0.225 \\
        & PP-StructureV3\cite{cui2025paddleocr} & - & 86.73 & 0.073 & 85.79 & 81.68 & 0.073 \\
    
        \midrule

        \multirow{6}{*}{\parbox{1.6cm}{\centering \textbf{General\\MLLMs}}}
        & GPT-4o\cite{gpt4o} & - & 75.02 & 0.217 & 79.70 & 67.07 & 0.148 \\
        & InternVL3\cite{internvl3} & 78B & 80.33 & 0.131 & 83.42 & 70.64 & 0.113 \\
        & InternVL3.5\cite{internvl35} & 241B & 82.67 & 0.142 & 87.23 & 75.00 & 0.125 \\
        & Qwen2.5-VL\cite{Qwen2.5-VL} & 72B & 87.02 & 0.094 & 88.27 & 82.15 & 0.102 \\
        & Gemini-2.5 Pro\cite{Gemini} & - & 88.03 & 0.075 & 85.82 & 85.71 & 0.097 \\
        & Qwen3-VL-235B\cite{qwen3vl2025} & 235B & 89.15 & 0.069 & 88.14 & 86.21 & 0.068 \\
        
        \midrule

        \multirow{4}{*}{\parbox{1.6cm}{\centering \textbf{Specialized\\MLLMs\\(Modular)}}}
        & Dolphin-1.5\cite{feng2025dolphin} & 0.3B & 83.21 & 0.092 & 80.78 & 78.06 & 0.124 \\
        & MonkeyOCR-pro-3B\cite{li2025monkeyocr} & 3B & 88.85 & 0.075 & 87.25 & 86.78 & 0.128 \\
        & MinerU2.5\cite{niu2025mineru25decoupledvisionlanguagemodel} & 1.2B & 90.67 & 0.047 & 88.46 & 88.22 & 0.044 \\
        & \underline{PaddleOCR-VL}\cite{cui2025paddlevl} & 0.9B & \underline{91.93} & \underline{0.039} & \underline{88.67} & \underline{91.01} & \underline{0.043} \\
        
        \midrule

        \multirow{5}{*}{\parbox{1.6cm}{\centering \textbf{Specialized\\MLLMs\\(End2End)}}}
        & Mistral OCR\cite{mistralocr} & - & 78.83 & 0.164 & 82.84 & 70.03 & 0.144 \\
        & POINTS-Reader\cite{liu2025points} & 3B & 80.98 & 0.134 & 79.20 & 77.13 & 0.145 \\
        & olmOCR\cite{olmocr} & 7B & 81.79 & 0.096 & 86.04 & 68.92 & 0.121 \\
        & Deepseek-OCR\cite{wei2025deepseek} & 3B & 87.01 & 0.073 & 83.37 & 84.97 & 0.086 \\
        & dots.ocr\cite{dots.ocr2025} & 3B & 88.41 & 0.048 & 83.22 & 86.78 & 0.053 \\

        \midrule
        \multirow{1}{*}{\parbox{1.6cm}{\centering \textbf{Ours}}}
        & \textbf{DocHumming} & 1B & \textbf{93.75} & \textbf{0.035} & \textbf{93.27} & \textbf{91.49} & \textbf{0.041} \\

        \bottomrule
    \end{tabular}
    \label{tab:DocBench_results}
    \vspace{-10pt}
\end{table*}

\paragraph{Training Data.}
\begin{itemize}
\item \textbf{Stage 1.} 
We train the model on 9 million atomic building blocks, paired with heterogeneous prompts to support element-level parsing.
\item \textbf{Stage 2.} 
We perform unified end-to-end training under homogeneous prompts by integrating \textit{DocMix-3M} with 1M structured instances sampled from atomic elements and 100K manually annotated real documents, predominantly from scanned/digital domains.

\end{itemize}

\paragraph{Evaluation Benchmarks and Metrics.}

\begin{itemize}
\item \textbf{OmniDocBench.}~\cite{ominibench} A benchmark of nine printed document types with full structural and reading-order annotations. We follow the evaluation protocol from~\cite{ominibench} for standard-layout assessment.

\item \textbf{XFUND.}~\cite{xfund} A multilingual printed-form benchmark with line-level annotations, covering six non-English/Chinese scripts; we follow the evaluation protocol in~\cite{bench_ccocr}.

\item \textbf{Wild-OmniDocBench.} 
A real-world captured variant of OmniDocBench with illumination and deformation artifacts for robustness evaluation; we follow the structured-output protocol of~\cite{ominibench} and use the degradation metrics of~\cite{wilddoc} against scanned/digital settings.

\end{itemize}

\paragraph{Baseline Settings.}

To facilitate a comprehensive evaluation, we group baseline models into three categories based on their methodological paradigms:

\begin{itemize}
    \item \textbf{Pipeline Tools.} Traditional modular systems where document parsing is decomposed into separate steps, such as layout analysis, OCR, and element classification.
    
    \item \textbf{General MLLMs.} MLLMs designed for broad vision-language tasks (e.g., QA, captioning) rather than specifically for structured document parsing.

    \item \textbf{Expert MLLMs.} Models tailored for document understanding, further categorized as 1) \textbf{E2E} models that directly map document images to structured outputs, and 2) \textbf{Modular} models that rely on layout decomposition and region-wise parsing under instruction-driven settings.
\end{itemize}

\noindent\textbf{Implementation Details.} We adopt InternVL2-1B\cite{mllm_internvl} as the base model and fine-tune all parameters across both training stages to obtain our document parsing model, \textbf{DocHumming}. In Stage 1, the model is trained for 2 epochs with a batch size of 512 and a learning rate of 4e-5. In Stage 2, training continues for 2 epochs with a reduced batch size of 256, a learning rate of 2e-5, and $\lambda=4$. A cosine learning rate decay is applied in both stages. The maximum output length is 8{,}192 tokens. All experiments are conducted on 16 NVIDIA H20 GPUs.


\subsection{Evaluation on Printed Document Parsing}

\noindent\textbf{Standardized Document Parsing.}  
We evaluate \textit{DocHumming} on standardized Chinese and English printed documents using the \textbf{OmniDocBench} benchmark.  
As shown in Table~\ref{tab:DocBench_results}, DocHumming consistently surpasses modular baselines in both full-document and element-level parsing accuracy.  
Built on \textbf{Realistic Scene Synthesis} and the \textbf{Document-Aware Training Recipe}, and trained with large-scale, diverse \textit{document-level} end-to-end supervision, DocHumming delivers robust, generalizable performance—validating the potential and practical feasibility of the end-to-end paradigm for document parsing.

\noindent\textbf{Multilingual Document Parsing.} We evaluate DocHumming’s multilingual parsing capabilities on the XFUND benchmark against other models supporting multilingual document parsing, with per-language results shown in Table \ref{tab:multilingual}. DocHumming demonstrates strong multilingual performance, benefiting from the multilingual supervision in DocMix-3M, which enhances its ability to handle diverse languages.

\begin{table}[h]
\vspace{-5pt}
\centering
\small
\caption{Performance comparison on the XFUND.}
\vspace{-10pt}
\aboverulesep = 0.1ex
    \belowrulesep = 0.2ex
\setlength{\tabcolsep}{5pt}
\resizebox{\linewidth}{!}{%
\begin{tabular}{lcccccc}
\toprule
Method & de & it & ja & es & pt    & fr    \\ \midrule
Mathpix~\cite{mathpix2025}              & \underline{83.90}   & 79.00    & 83.95    & 81.23   & 76.8  & 73.29 \\
GOT-OCR\cite{mllm_got}              &  83.45  &  47.84  &  50.81   &  64.60  & 69.12 & 39.09 \\
MistralOCR              & 78.06  & 63.21   & 63.61    & 60.09   & 58.89 & 59.83 \\
Dots.ocr            & 53.94 & 70.15 & 70.73 & 66.87 & 60.75 & 59.91 \\
GPT-4o                   & 78.94  & 76.83   & 83.83    & 80.37   & 77.25 & 76.08 \\
Qwen3-VL-235B            & 74.45  & 78.36   & 85.30    & 76.36   & 76.73 & 69.73 \\
MinerU2.5            &  83.27 &  76.57 & 82.37 & 78.59 & 76.30 & 70.95 \\
PaddleOCR-VL            &  80.98  &  75.93  &  85.82  & 80.48 & 77.68 & 72.46 \\
Gemini2.5-Pro                   & 82.43  &  \underline{79.22}  &  \underline{86.32}  &  \underline{81.41}  &\underline{80.50} & \underline{76.69} \\

\midrule
\textbf{DocHumming}       & \textbf{85.15}  & \textbf{80.06}   & \textbf{87.99}    & \textbf{84.39}   & \textbf{83.67} & \textbf{77.48} \\ \bottomrule
\end{tabular}
}
\vspace{-6pt}

\label{tab:multilingual}
\end{table}


\begin{table*}[ht]
\vspace{-10pt}
\centering
\small
\caption{Performance comparison on the Wild-OmniDocBench.}
\vspace{-10pt}
\resizebox{\textwidth}{!}{
    \begin{tabular}{clcccccccc}
        \toprule
        \multirow{2}{*}{\textbf{Model$_{\mathrm{Type}}$}} & \multirow{2}{*}{\textbf{Model}} 
        & \multicolumn{2}{c}{\textbf{Overall}} 
        & \multicolumn{2}{c}{\textbf{Text$^{\mathrm{1\text{-}Edit}}$ $\uparrow$}} 
        & \multicolumn{2}{c}{\textbf{Formula$^{\mathrm{CDM}}$ $\uparrow$}} 
        & \multicolumn{2}{c}{\textbf{Table$^{\mathrm{TEDS}}$ $\uparrow$}} \\ 
        \cmidrule(l){3-4} \cmidrule(l){5-6} \cmidrule(l){7-8} \cmidrule(l){9-10}
        & & \textbf{Origin} & \textbf{Wild} & \textbf{Origin} & \textbf{Wild} & \textbf{Origin} & \textbf{Wild} & \textbf{Origin} & \textbf{Wild} \\
        \midrule

        \multirow{1}{*}{\parbox{1.6cm}{\centering \textbf{General}}}
        & Qwen3-VL-235B 
        & 89.15 
        & \underline{79.69}$_{\decrease{-9.46}}$ 
        & 93.1 
        & \underline{90.1}$_{\decrease{-3.0}}$ 
        & 88.14 
        & \underline{80.67}$_{\decrease{-7.47}}$ 
        & 86.21 
        & 68.31$_{\decrease{-17.90}}$ \\
        \midrule

        \multirow{3}{*}{\parbox{1.6cm}{\centering \textbf{Modular}}}
        & MonkeyOCR-pro-3B 
        & 88.85 
        & 70.00$_{\decrease{-18.85}}$ 
        & 92.5 
        & 78.9$_{\decrease{-13.6}}$ 
        & 87.25 
        & 63.27$_{\decrease{-23.98}}$ 
        & 86.78 
        & 67.83$_{\decrease{-18.95}}$ \\
        & MinerU2.5 
        & 90.67 
        & 70.91$_{\decrease{-19.76}}$ 
        & 95.3 
        & 78.2$_{\decrease{-17.1}}$ 
        & 88.46 
        & 64.37$_{\decrease{-24.09}}$ 
        & 88.22 
        & 70.15$_{\decrease{-18.07}}$ \\
        & PPOCR-VL 
        & \underline{91.93} 
        & 72.19$_{\decrease{-19.74}}$ 
        & \underline{96.1} 
        & 76.8$_{\decrease{-19.3}}$ 
        & \underline{88.67} 
        & 65.54$_{\decrease{-23.13}}$ 
        & \underline{91.01} 
        & \underline{74.24}$_{\decrease{-16.77}}$ \\
        
        \midrule

        \multirow{3}{*}{\parbox{1.6cm}{\centering \textbf{End2End}}}
        & DeepSeek-OCR 
        & 87.01 
        & 74.23$_{\decrease{-12.78}}$ 
        & 92.7 
        & 82.2$_{\decrease{-10.5}}$ 
        & 83.37 
        & 70.07$_{\decrease{-13.30}}$ 
        & 84.97 
        & 70.41$_{\decrease{-14.56}}$ \\
        & dots.ocr 
        & 88.41 
        & 78.01$_{\decrease{-10.40}}$ 
        & 95.2 
        & 87.9$_{\decrease{-7.3}}$ 
        & 83.22 
        & 74.23$_{\decrease{-8.99}}$ 
        & 86.78 
        & 71.89$_{\decrease{-14.89}}$ \\
        & DocHumming 
        & \textbf{93.75} 
        & \textbf{87.03}$_{\decrease{-6.72}}$ 
        & \textbf{96.5} 
        & \textbf{93.1}$_{\decrease{-3.4}}$ 
        & \textbf{93.27} 
        & \textbf{83.25}$_{\decrease{-10.02}}$ 
        & \textbf{91.49} 
        & \textbf{84.72}$_{\decrease{-6.77}}$ \\
        \bottomrule
    \end{tabular}
}
\vspace{-6pt}
\label{tab:docml}
\end{table*}

We evaluate several competitive models from each methodological paradigm on \textbf{Wild-OmniDocBench} to assess robustness under real-world capture.
As shown in Table~\ref{tab:docml}, \textbf{DocHumming} achieves the highest overall performance, demonstrating strong resilience to illumination variations, geometric distortions, and background interference present in naturally captured documents.

Notably, when compared to the original OmniDocBench results, end-to-end parsing models exhibit significantly less performance degradation than cascaded counterparts.  
This observation supports our hypothesis that real-world captured conditions introduce substantial challenges to layout analysis—on which modular pipelines heavily rely—while end-to-end approaches, free from explicit segmentation dependency, maintain more stable and accurate parsing.  
These findings highlight the practicality of the end-to-end paradigm as a robust and deployment-ready solution for real-world document understanding.

\subsection{Ablation Study}

To validate our approach, we conduct ablation studies on both the \textbf{Realistic Scene Synthesis} (RSS) and the \textbf{Document-Aware Training Recipe} (DATR). Experiments are performed on \textit{OmniDocBench} and \textit{Wild-OmniDocBench}, representing printed and real-world scenarios. Additionally, we introduce a \textbf{repetition rate} metric, defined as the fraction of outputs that (i) contain an identical structured pattern repeated more than 10 times and (ii) reach the maximum generation length, to assess the stability of structured decoding. Results are shown in Table~\ref{tab:ablation_extended_results}.

\begin{table}
    \centering
    \small
    \caption{Extended ablation study on Realistic Scene Synthesis. and Document-Aware Training Recipe. }
    \vspace{-10pt}
    \setlength{\tabcolsep}{3pt}
    \begin{tabular}{cccccccc}
        \toprule
        \multirow{2}{*}{\#} & \multirow{2}{*}{\textbf{RSS}} & \multicolumn{2}{c}{\textbf{DATR}} &\multicolumn{2}{c}{\textbf{OmniDocBench}} & \multicolumn{2}{c}{\textbf{Wild-OmniDocBench}}  \\
        \cmidrule(l){3-4} \cmidrule(l){5-6}  \cmidrule(l){7-8} 
        & & \textbf{ST} & \textbf{PTP} & \textbf{Overall}$\mathrm{_{\uparrow}}$ & \textbf{Repeat}$\mathrm{_{\downarrow}}$ & \textbf{Overall}$\mathrm{_{\uparrow}}$ & \textbf{Repeat}$\mathrm{_{\downarrow}}$ \\
        \midrule
        1 & $\times$ & $\checkmark$ & $\checkmark$ & 89.96 & \enspace 4.7 & 78.82 & 8.6\\
        2 & $\checkmark$ & $\times$ & $\checkmark$ & 88.74 & 4.6 & 84.90 & 5.4\\
        3 & $\checkmark$ & $\checkmark$ & $\times$ & 91.24 & 4.2 & 85.39 & 4.9\\
        4 & $\checkmark$ & $\checkmark$ &$\checkmark$ &  \textbf{93.75} &  \enspace \textbf{2.1} & \textbf{87.03} & \textbf{4.3}\\
        
        \bottomrule
    \end{tabular}
    \vspace{-10pt}
    \label{tab:ablation_extended_results}
\end{table}

\noindent\textbf{Realistic Scene Synthesis.}
We build a size- and language-matched baseline using a conventional PDF-to-LaTeX pipeline~\cite{mllm_got} to isolate the effect of our synthesis strategy.
Comparing \#1 with \#4, our method delivers clear gains on both settings: +3.79 Overall on OmniDocBench and +8.21 on Wild-OmniDocBench, while the repetition rate drops from 4.7$\!\rightarrow\!$2.1 and 8.6$\!\rightarrow\!$4.3, respectively.
These improvements, evident in both wild and printed settings, indicate that RSS provides artifact-aware and diversity-rich supervision. By covering varied layouts, reading orders, and visual conditions, RSS strengthens end-to-end parsing beyond clean digital data.

\noindent\textbf{Document-Aware Training Recipe.}
We ablate the \textit{Structure-Token Aware Optimization} (ST) and the \textit{Progressive Training Paradigm} (PTP).

Removing ST while keeping RSS and PTP fixed (compare \#2 vs.\ \#4), with identical data, schedule, and hyperparameters, reduces Overall by 5.01 on OmniDocBench and 2.13 on Wild-OmniDocBench, and increases repetition from 2.1$\rightarrow$4.6 and 4.3$\rightarrow$5.4, respectively. 
This confirms that emphasizing structurally critical tokens stabilizes decoding and improves structural fidelity. 

Removing PTP, implemented by collapsing the two-stage curriculum (merging Stage~1 and Stage~2 data) and training end-to-end with a fixed learning rate of 4e-5 while keeping RSS and ST unchanged (compare \#3 vs.\ \#4), reduces Overall by 2.51 / 1.64 on OmniDocBench / Wild-OmniDocBench and increases repetition from 2.1$\rightarrow$4.2 and 4.3$\rightarrow$4.9, respectively. This aligns with the LLM curriculum prior that progressing from short- to long-context learning stabilizes optimization and improves long-context consistency.

Overall, combining RSS with ST and PTP (\#4) achieves the best accuracy–stability trade-off across printed and real-world settings, validating the effectiveness of our data–training co-design for robust end-to-end document parsing.

\begin{table}
    \centering
    \small
    \caption{Extended ablation study on Realistic Scene Synthesis. and Document-Aware Training Recipe. }
    \vspace{-10pt}
    \setlength{\tabcolsep}{3pt}
    \begin{tabular}{cccccccc}
        \toprule
        \multirow{2}{*}{\#} & \multirow{2}{*}{\textbf{Data setting}} & \multicolumn{2}{c}{\textbf{OmniDocBench}} & \multicolumn{2}{c}{\textbf{Wild-OmniDocBench}}  \\
        \cmidrule(l){3-4} \cmidrule(l){5-6}  
        & & \textbf{Overall}$\mathrm{_{\uparrow}}$ & \textbf{Repeat}$\mathrm{_{\downarrow}}$ & \textbf{Overall}$\mathrm{_{\uparrow}}$ & \textbf{Repeat}$\mathrm{_{\downarrow}}$ \\
        \midrule
        1 & Manual-100k & 89.26 & \enspace 5.4 & 80.20 & 7.8\\
        2 & DocMix-1M & 85.41 & 7.9 & 76.66 & 8.9\\
        3 & DocMix-2M & 88.14 & 5.6 & 80.08 & 6.2\\
        4 & DocMix-3M &  \textbf{89.96} &  \enspace 3.8 & 83.21 & \textbf{4.8}\\
        5 & DocMix-4M &  89.31 &  \enspace \textbf{3.7} & \textbf{83.52} & 4.9\\

        \bottomrule
    \end{tabular}
    \label{tab:ablation_data_scall}
    \vspace{-10pt}
\end{table}

\subsection{Data Benefits and Scaling Law}
\label{subsec:data_scaling}

To quantify the effect of supervision sources and data scale in Stage~2, we compare training with only \textbf{manual} annotations (100K scanned/digital pages) against training with only \textbf{RSS} synthetic pages at different scales (DocMix-1M/2M/3M/4M), keeping Stage~1 and the training recipe unchanged. Results are reported in Table~\ref{tab:ablation_data_scall}.

\noindent\textbf{Synthetic data benefits.}
At small scale (DocMix-1M), synthetic supervision underperforms manual data on both benchmarks (85.41/76.66 vs.\ 89.26/80.20) and shows higher repetition. 
As scale increases, performance improves steadily: DocMix-2M closes most of the gap, and DocMix-3M surpasses manual supervision on both \textit{OmniDocBench} and \textit{Wild-OmniDocBench} while substantially reducing repetition (e.g., 89.96 vs.\ 89.26 and 3.8 vs.\ 5.4; 83.21 vs.\ 80.20 and 4.8 vs.\ 7.8). 
These trends substantiate a data scaling law realized via RSS, yielding tangible gains for end-to-end document parsing while remaining readily extensible and promising to mitigate the high cost of curating document-level end-to-end supervision.

\noindent\textbf{Scaling law and saturation.}
We observe clear gains from 1M$\rightarrow$2M and 2M$\rightarrow$3M, followed by \emph{diminishing returns} beyond 3M: DocMix-4M yields marginal changes (slight improvement on Wild overall, small fluctuation in repetition), indicating a near-saturation regime around 3M. 
This suggests that scaling benefits are ultimately bounded by the \emph{capacity of the bottom-up generators}: with a fixed element repository and template pool, additional samples increasingly recombine similar primitives, limiting the introduction of genuinely novel structures and visual conditions.

\subsection{Analyzing Repetitive Decoding in End-to-End Parsing}
\label{subsec:repetition_analysis}

We analyze the causes of repetitive predictions and attribute the repetition rate to three factors:

\noindent\textbf{(1) Stability of structured decoding.}
Autoregressive decoding is fragile on structured outputs (tables, forms, lists). 
Uniform token-level training underweights structural tokens that require strict consistency, leading to unstable boundary decoding and repetition. 
Our \textit{Structure-Token Aware Optimization} (ST) emphasizes these tokens and consistently lowers repetition (cf. \#2 vs.\ \#4 in Table~\ref{tab:ablation_extended_results}).

\noindent\textbf{(2) Train–test distribution mismatch.}
With identical prompts, models overfit structural priors of the training domain. 
Rule-based PDF-to-\LaTeX{} pipelines provide limited layout templates, hurting generalization and triggering repetition out of distribution. 
\textit{Realistic Scene Synthesis} (RSS) broadens layout variants, reading orders, and capture-aware augmentations, acting as a regularizer and lowering repetition in both printed and wild scenarios (cf. \#1 vs.\ \#4 in Table~\ref{tab:ablation_extended_results}).

\noindent\textbf{(3) Training stability.}
Page-level document parsing entails long-context training, and unstable convergence can increase repetition.
Our \textit{Progressive Training Paradigm} (PTP) transitions from element-level to page-level supervision, stabilizing optimization and alleviating repetition (cf. \#3 vs.\ \#4 in Table~\ref{tab:ablation_extended_results}).

\noindent\textbf{(4) Data scale.}
Insufficient end-to-end supervision increases repetition. Expanding page-level data via RSS (cf. DocMix-1M$\rightarrow$3M in Table~\ref{tab:ablation_data_scall}) reduces repetition and improves overall accuracy; gains taper near 3M as element/template diversity becomes the limiting factor.

\subsection{Limitations and Future Work}
\label{subsec:limitations}

Despite the gains from our \textbf{data–training co-design framework}, fully end-to-end document parsing remains imperfect. \textit{DocHumming}—as well as contemporaneous E2E models—shows the following limitations:

\begin{itemize}
\item \textbf{Irregular, interleaved layouts.} Performance drops on highly non-standard pages where text blocks interleave or nest (e.g., newspapers, posters), making reading order and structural boundaries ambiguous.

\item \textbf{Ultra–high-resolution pages.} Input-resolution limits force downsampling or tiling on very large pages, which may induce repeated or missing content in long tables, dense formulas, or multi-column layouts.

\item \textbf{Computational efficiency.} Benefiting from its 1B-parameter scale, DocHumming achieves higher through put than recent end-to-end parsers (e.g., DeepSeek-OCR), 
yet parsing text-dense pages still takes about $\sim$ 3 s per 
page, limiting interactive use.

\end{itemize}

\noindent\textbf{Future work.}
On the \textbf{model} side, we will enhance layout awareness for irregular structures, adopt resolution-adaptive modeling for ultra–large pages with cross-tile consistency, and reduce latency with lighter backbones and efficient decoding. 
On the \textbf{data} side, building on \textit{Realistic Scene Synthesis}, we will synthesize page-level content with \emph{semantic and logical coherence across elements} to better supervise holistic understanding. 



\section{Conclusion}

In this paper, we present \textbf{DocHumming}, an end-to-end document parsing framework that combines \textbf{Realistic Scene Synthesis} for scalable data generation and a \textbf{Document-Aware Training Recipe} for effective training. Our approach shows strong performance on multilingual and real-world datasets, demonstrating the ability of end-to-end models to surpass modular pipelines. With \textbf{Wild-OmniDocBench}, we offer a benchmark to evaluate robustness across diverse real-world scenarios. Our results highlight the advantages of unified document parsing, advancing practical and scalable document understanding for real-world applications.
\section*{Acknowledgements} 

This work is supported by the National Natural Science Foundation of China (Grant NO 62376266 and 62406318).
{
    \small
    \bibliographystyle{ieeenat_fullname}
    \bibliography{main}
}


\end{document}